\documentclass{article}
\usepackage{latinxai}
\usepackage[utf8]{inputenc}
\usepackage{graphicx}
\usepackage{amsmath}

\title{Y-GAN: A Generative Adversarial Network for Depthmap Estimation from Multi-camera Stereo Images}
\author{
  Miguel Alonso Jr. \\
  School of Computing and Information Science\\
  Florida International University University\\
  Miami, FL 33199 \\
  \texttt{malonsoj@cs.fiu.edu} \\
}

\begin{document}

\maketitle

\section{Introduction}
Depth perception is a key component for autonomous systems that interact in the real world, such as delivery robots, warehouse robots, and self-driving cars. Tasks in autonomous robotics such as 3D object recognition, simultaneous localization and mapping (SLAM), path planning and navigation, require some form of 3D spatial information \cite{Ayache1991TrinocularRobotics}. Depth perception is a long standing research problem in computer vision and robotics, and has had a long history. Many approaches using deep learning, ranging from structure from motion, shape-from-X, monocular, binocular, and multi-view stereo, have yielded acceptable results \cite{Godard2017UnsupervisedConsistency,Ummenhofer2017DeMoN:Stereo, Laina2016DeeperNetworks}. However, there are several shortcomings of these methods such as requiring expensive hardware, needing supervised training data, no ground truth data for comparison, and disregard for occlusion \cite{Godard2017UnsupervisedConsistency}. In order to address these shortcomings, this work proposes a new deep convolutional generative adversarial network architecture, called Y-GAN, that uses data from three cameras to estimate a depth map for each frame in a multi-camera video stream.

\section{Research Problem}
Generative Adversarial Networks (GAN), specifically, Deep Convolutional Generative Adversarial Networks (DCGAN), have gained tremendous popularity since being first introduced in 2014, due to their unique ability to generate realistic images at random from learned distributions in the training data \cite{Goodfellow2014GenerativeNetworks,Radford2015UnsupervisedNetworks}. Since depth map estimation from stereo or monocular images can be considered a image generation problem, GANs or DCGANs are well suited \cite{Almalioglu2018GANVO:Networks,Gwn2018GenerativeVideo,Aleotti2019GenerativePrediction,Chen2018RethinkingTraining,Pilzer2018UnsupervisedNetworks}. Thus, the main research problem in this proposed work in progress is to develop a deep learning architecture that takes as input three stereo images arranged in a left-center-right orientation and produce the depth map that corresponds to the center image. An adjacent, but related research problem  also considered in this work is to structure this deep learning architecture in such a way as to be able to train it in an unsupervised or semi-supervised fashion.

\section{Motivation}
The motivation behind taking a purely vision based approach to autonomous robotics is that most modern autonomous robots, such as self-driving cars, rely on very expensive equipment, such as LiDAR, RADAR, and GPS \footnote{https://automotivelectronics.com/cost-of-components-of-a-self-driving-car/}, and those costs will not be going down anytime soon. A purely vision based system currently costs substantially less. Additionally, focusing on purely vision based deep learning approaches that mimic the way in which humans perceive depth and localize themselves in the world can lead to a greater understanding of the underlying mechanisms of perception and cognition in the human brain. 

\section{Contribution}
The main contribution of this work, if successful once completed, is to yield a new  generative adversarial network architecture that can accurately estimate depth from a monocular image. The following two sections describe the proposed network architecture and next steps.

\subsection{Y-GAN Network Architecture}
\begin{figure}[t!]
    \centering
    \includegraphics[scale=0.5]{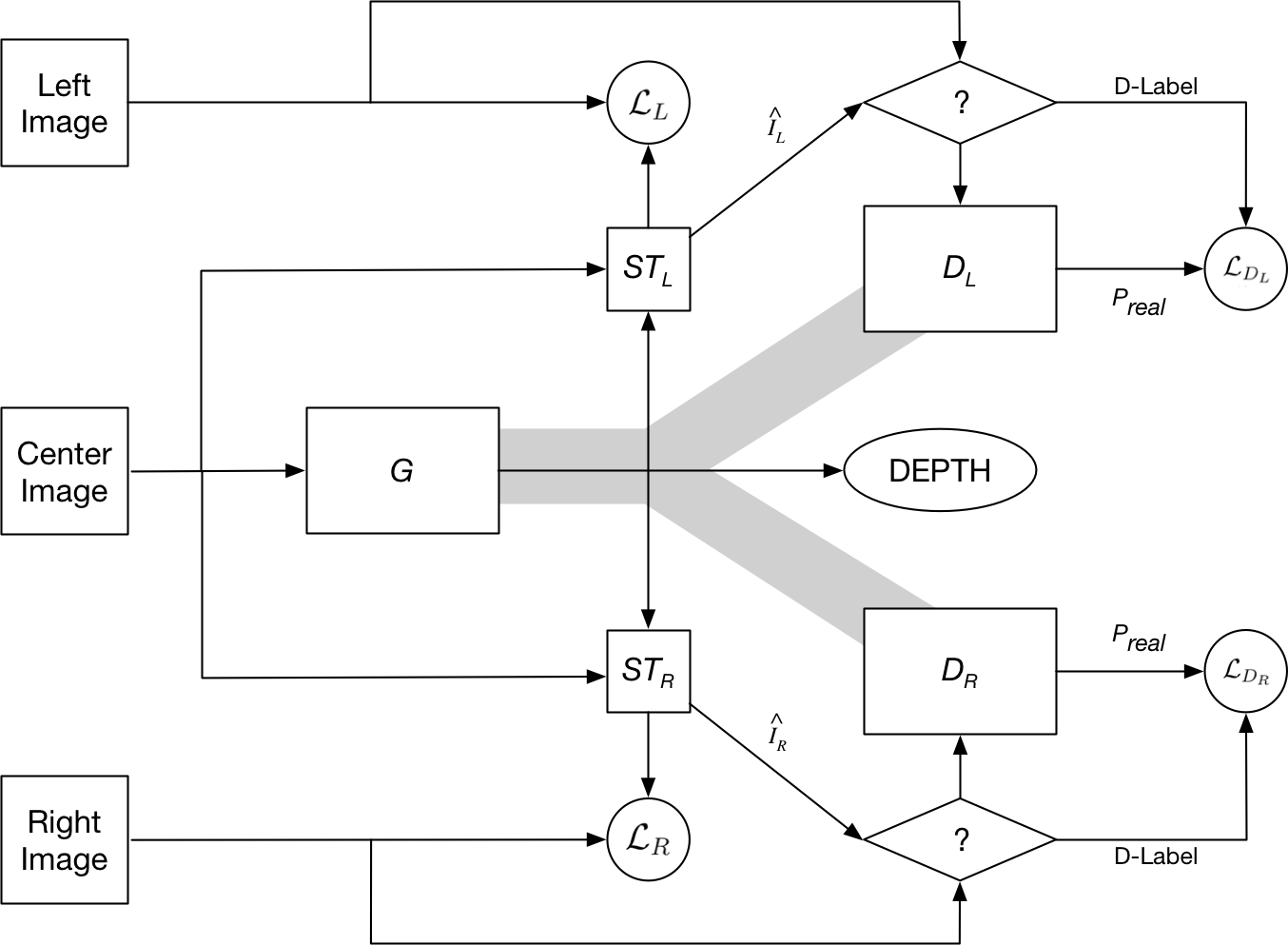}
    \caption{Y-GAN Architecture}
    \label{fig:ygan}
\end{figure}

Figure \ref{fig:ygan} shows the structure of the Y-GAN. This GAN architecture has a single generator, $G$, and two discriminators, $D_L$, the left image discriminator, and $D_R$, the right image discriminator. $G$ is a U-NET with the left half of the U being a pre-trained RESNET34 \cite{He2015DeepRecognition}. The right half of the U-NET is not frozen and is trained along with the rest of the Y-GAN. The goal of the generator is to generate a depthmap from the center image. This depthmap is then used along with a spatial transformer \cite{Jaderberg2015SpatialNetworks} ($ST_L$ and $ST_R$) and the center image to generate a reconstruction of the left and right images ($\hat{I_L}$ and $\hat{I_R}$). $\hat{I_L}$ and $\hat{I_R}$ are sent to the discriminator for judgement. The GAN loss functions that will be used to train the network are described in section \ref{lossfunctions}. 
\subsection{Loss Functions} \label{lossfunctions}
Similar to GANs, the loss functions $\mathcal{L}_{D_L}$ and $\mathcal{L}_{D_R}$ for $D_L$ and $D_R$ are $logD_L(x)$ and $logD_R(x)$, respectively. The goal of training the discriminators is to minimize this loss. However, in contrast to GANs, a YGAN would need a very different loss function to train the generator, $G$. The loss function for the generator is given by \begin{equation}
    \mathcal{L}_G = log(1-D_L(\hat{I_L})) + log(1-D_R(\hat{I_R})) + \mathcal{L}_L + \mathcal{L}_R
\end{equation}
where $\mathcal{L}_L$ and $\mathcal{L}_R$ are given by
\begin{equation}
    \mathcal{L}_K = \frac{1}{N}\sum_N [||I_K - \hat{I_K}|| + SSIM(I_K, \hat{I_K})]
\end{equation}
and $SSIM(I_K, \hat{I_K})$ is a structural similarity measure \cite{Godard2017UnsupervisedConsistency}. Including $\mathcal{L}_L$ and $\mathcal{L}_R$ as constraints ensures that the reconstructed images match the ground truth, thereby biasing the generator to produce an accurate depthmap for the center image.
\section{Next Steps}
The next steps for this work include synthetic dataset generation (including ground truth depthmaps for estimating accuracy), training and testing on the synthetic data, and transfer learning evaluation on a real multi-camera dataset.

\newpage
\bibliographystyle{ieeetr}
\bibliography{references}

\begin{thebibliography}{10}

\bibitem{Ayache1991TrinocularRobotics}
N.~Ayache and F.~Lustman, ``{Trinocular stereo vision for robotics},'' {\em
  IEEE Transactions on Pattern Analysis and Machine Intelligence}, vol.~13,
  no.~1, pp.~73--85, 1991.

\bibitem{Godard2017UnsupervisedConsistency}
C.~Godard, O.~Mac~Aodha, and G.~J. Brostow, ``{Unsupervised monocular depth
  estimation with left-right consistency},'' in {\em Proceedings - 30th IEEE
  Conference on Computer Vision and Pattern Recognition, CVPR 2017},
  vol.~2017-Janua, pp.~6602--6611, IEEE, 7 2017.

\bibitem{Ummenhofer2017DeMoN:Stereo}
B.~Ummenhofer, H.~Zhou, J.~Uhrig, N.~Mayer, E.~Ilg, A.~Dosovitskiy, and
  T.~Brox, ``{DeMoN: Depth and motion network for learning monocular stereo},''
  in {\em Proceedings - 30th IEEE Conference on Computer Vision and Pattern
  Recognition, CVPR 2017}, vol.~2017-Janua, pp.~5622--5631, IEEE, 7 2017.

\bibitem{Laina2016DeeperNetworks}
I.~Laina, C.~Rupprecht, V.~Belagiannis, F.~Tombari, and N.~Navab, ``{Deeper
  depth prediction with fully convolutional residual networks},'' in {\em
  Proceedings - 2016 4th International Conference on 3D Vision, 3DV 2016},
  pp.~239--248, IEEE, 10 2016.

\bibitem{Goodfellow2014GenerativeNetworks}
I.~J. Goodfellow, J.~Pouget-Abadie, M.~Mirza, B.~Xu, D.~Warde-Farley, S.~Ozair,
  A.~Courville, and Y.~Bengio, ``{Generative Adversarial Networks},'' 6 2014.

\bibitem{Radford2015UnsupervisedNetworks}
A.~Radford, L.~Metz, and S.~Chintala, ``{Unsupervised Representation Learning
  with Deep Convolutional Generative Adversarial Networks},'' 11 2015.

\bibitem{Almalioglu2018GANVO:Networks}
Y.~Almalioglu, M.~R.~U. Saputra, P.~P.~B. de~Gusmao, A.~Markham, and
  N.~Trigoni, ``{GANVO: Unsupervised Deep Monocular Visual Odometry and Depth
  Estimation with Generative Adversarial Networks},'' 9 2018.

\bibitem{Gwn2018GenerativeVideo}
K.~Gwn, K.~Reddy, M.~Giering, and E.~A. Bernal, ``{Generative adversarial
  networks for depth map estimation from RGB video},'' in {\em IEEE Computer
  Society Conference on Computer Vision and Pattern Recognition Workshops},
  vol.~2018-June, pp.~1258--1266, 2018.

\bibitem{Aleotti2019GenerativePrediction}
F.~Aleotti, F.~Tosi, M.~Poggi, and S.~Mattoccia, ``{Generative Adversarial
  Networks for Unsupervised Monocular Depth Prediction},'' pp.~337--354, 2019.

\bibitem{Chen2018RethinkingTraining}
R.~Chen, F.~Mahmood, A.~Yuille, and N.~J. Durr, ``{Rethinking Monocular Depth
  Estimation with Adversarial Training},'' 8 2018.

\bibitem{Pilzer2018UnsupervisedNetworks}
A.~Pilzer, D.~Xu, M.~Puscas, E.~Ricci, and N.~Sebe, ``{Unsupervised adversarial
  depth estimation using cycled generative networks},'' in {\em Proceedings -
  2018 International Conference on 3D Vision, 3DV 2018}, pp.~587--595, 7 2018.

\bibitem{He2015DeepRecognition}
K.~He, X.~Zhang, S.~Ren, and J.~Sun, ``{Deep Residual Learning for Image
  Recognition},'' 12 2015.

\bibitem{Jaderberg2015SpatialNetworks}
M.~Jaderberg, K.~Simonyan, A.~Zisserman, and K.~Kavukcuoglu, ``{Spatial
  Transformer Networks},'' 6 2015.

\end{thebibliography}

\end{document}